\documentclass[letterpaper, 10pt, conference]{ieeeconf}  % Comment this line out if you need a4paper

\IEEEoverridecommandlockouts                              % This command is only needed if 
\usepackage{dblfloatfix}
\usepackage{graphicx} % for pdf, bitmapped graphics files
\usepackage{amsmath} % assumes amsmath package installed
\usepackage{amssymb}  % assumes amsmath package installed
\usepackage{kantlipsum}
\usepackage{multirow}
\usepackage{cite}
\usepackage[normalem]{ulem} 
\usepackage{mathtools}
\newcommand{\defeq}{\vcentcolon=}
\begin{document}
\title{\Large \bf Diverse Critical Interaction Generation for Planning and Planner Evaluation}
\author{Zhao-Heng Yin$^{*,1}$, Lingfeng Sun$^{*,2}$, Liting Sun$^2$, Masayoshi Tomizuka$^2$, Wei Zhan$^2$ % <-this % stops a space
\thanks{* Equal contribution.}% <-this % stops a space
\thanks{$^{1}$Zhao-Heng Yin is with the Department of Computer Science and Technology, Nanjing University, Nanjing 210023, Jiangsu, PRC. \texttt{zhaohengyin@gmail.com}}%
\thanks{$^{2}$ Lingfeng Sun,  Liting Sun, Masayoshi Tomizuka, Wei Zhan are with the Department of Mechanical Engineering, University of California, Berkeley, Berkeley, CA 94720, USA. 	\texttt{\{lingfengsun, litingsun, tomizuka, wzhan\}@berkeley.edu}}%
}

\maketitle
%\thispagestyle{empty}
%\pagestyle{empty}

%%%%%%%%%%%%%%%%%%%%%%%%%%%%%%%%%%%%%%%%%%%%%%%%%%%%%%%%%%%%%%%%%%%%%%%%%%%%%%%%
\begin{abstract}
 %Route planner is a crucial component in the decision process of self-driving cars, which outputs several future way points for the controlled vehicle. In order to ensure that the designed trajectory planners are robust and will make safe control decisions across various scenarios, one needs to design comprehensive testing cases to test the designed planning algorithm. However, most existing testing methods design safe-critical interactions by hand engineering, which requires much human effort and can become quite hard and tricky in some complex scenarios. Moreover, we notice that existing test methods usually focus on normal~(safe) scenarios. But what is more important is that how these planning algorithms will react to safe-critical cases, when the motion of its surrounding vehicle is likely to result in accidents. 
 Generating diverse and comprehensive interacting agents to evaluate the decision-making modules is essential for the safe and robust planning of autonomous vehicles~(AV). Due to efficiency and safety concerns, most researchers choose to train interactive adversary~(competitive or weakly competitive) agents in simulators and generate test cases to interact with evaluated AVs. However, most existing methods fail to provide both natural and critical interaction behaviors in various traffic scenarios. To tackle this problem, we propose a styled generative model RouteGAN that generates diverse interactions by controlling the vehicles separately with desired styles. By altering its style coefficients, the model can generate trajectories with different safety levels serve as an online planner. Experiments show that our model can generate diverse interactions in various scenarios. We evaluate different planners with our model by testing their collision rate in interaction with RouteGAN planners of multiple critical levels.
 % Researchers have proposed to generate interaction data by deep generative models. However, one restriction of them is that they can only generate the trajectories of all vehicles involved in the interaction simultaneously. In some cases we require that the trajectory of some particular vehicle is assigned by other planners. In other words, the problem is how to . In this paper, we proposed an algorithm.
\end{abstract}
\section{Introduction}
Self-driving vehicles are expected to make transportation systems much more efficient in the future with smart decision-making systems that can avoid irrational behaviors leading to potential dangers. The safety and robustness evaluation of autonomous vehicle planners during online operation remains an essential but unsolved problem. As in all the other engineering fields, one fundamental philosophy to tackle this problem is through comprehensive testing. In reality, it takes hundreds of miles for autonomous vehicles to encounter various safe-critical cases. Such an evaluation process is both time-consuming and risky since we cannot control other traffic participants' behavior on the road. As a result, researchers have proposed multiple evaluation methods in simulator environments, where the key problem degenerates to designing diverse and natural test cases. To take advantage of prior knowledge and collected natural interaction data, researchers propose data-driven and learning methods for planner evaluation. In the testing process, \textit{learned} interactive adversary~(competitive or weakly competitive) agents are controlled to interact with the tested vehicle, and we use safety metrics like collision rate to evaluate planners' performance. By varying the environment and adversary agents in test cases, we can figure out possible failure modes and improve the decision-making algorithms. A common approach in previous test case generation methods is to sample diverse initial states of the adversary agents~\cite{testing_gen_whd, learn_to_collide}. However, one drawback of this approach is that they usually assume over-simplified adversary agents with little reaction to the tested vehicle. In reality, the adversary agents usually alter the speed and orientation based on observation of other vehicles during the interaction. Another popular approach is to train adversary agents with Reinforcement Learning~(RL) to minimize the driving performance of tested agent~\cite{test_gen_rl}. The trained adversary agents have complex reacting driving behavior, but their reactions are usually not natural since they do not learn from human driving data. Moreover, they cannot be generalized to different road structures. To summarize, our main question is:
\begin{center}
    \textit{Can we design interactive adversary agents for testing, which have \uline{diverse}, \uline{natural} behaviors in \uline{various scenarios}?}
\end{center}

We notice that some previous data-driven trajectory generation methods can produce diverse, near-authentic trajectories~\cite{mtg, cmts}. However, the purpose of these methods is to generate the whole interaction data. In other words, they generate the trajectory of all the agents jointly, rather than controlling a single agent conditioning on its observation on other agents. Therefore, we cannot directly use joint trajectories generation to control adversarial agents for testing case generation. Our proposed method is designed to remove such restrictions and apply data-driven methods to adversary agents training in planner evaluation. 

\begin{figure}[t]
    \centering
    \includegraphics[width=0.5\textwidth]{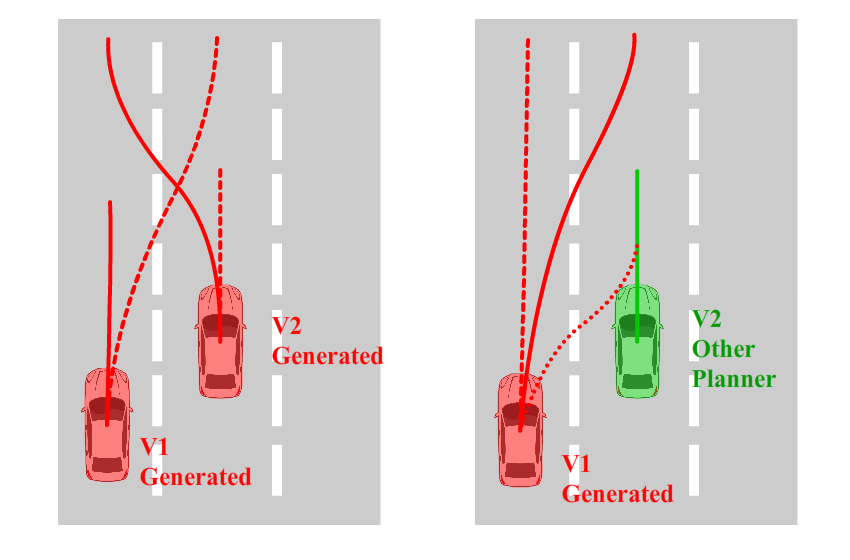}
    \caption{\textbf{Left:} Previous generative methods produce various interactions by generating the trajectory of all the involved vehicles. \textbf{Right:} We plan on a single vehicle to generate diverse interactions and allow the other vehicle to use arbitrary planners~(including the proposed RouteGAN).}
    \label{fig:idea}
\end{figure}
In this paper, we propose RouteGAN, a deep generative model that generates diverse interactive behavior of a controlled vehicle using observations of the surrounding vehicles. Instead of generating all the trajectories jointly, RouteGAN controls a single vehicle and allows controlling other vehicles with RouteGAN or any other planning modules as shown in Figure~\ref{fig:idea}. To ensure the diversity of generated trajectories, we use a style variable to control the proposed generation process. In particular, one dimension of the style variable represents how critical the generated trajectory is, which allows us to produce safe, near-critical, and critical interactions with the surrounding vehicles. As a result, RouteGAN can be used as a safe planner during planning and planner for adversary agents during planner evaluation.

Our contribution can be summarized as follows.
\begin{itemize}
    \item We propose RouteGAN, which can produce styled behavior of a single agent in multi-agent interaction scenarios. The proposed model controls the styles of agents separately and iteratively generates the whole interactions.
    \item Multi-branch safe/critical discriminators and Auxiliary Distribution Network are designed to ensure style control over generated behavior of the interacting vehicle. Experiments show that the model can generate diverse interactions in various scenarios.
    \item We use RouteGAN as an online opponent vehicle planner to test the performance of different planners. Results show that varying style input of RouteGAN controlled opponent vehicle increases the collision rate for rule-based and data-driven planning models.
\end{itemize}
\begin{figure*}[t]
    \centering
    \includegraphics[width=0.98\textwidth]{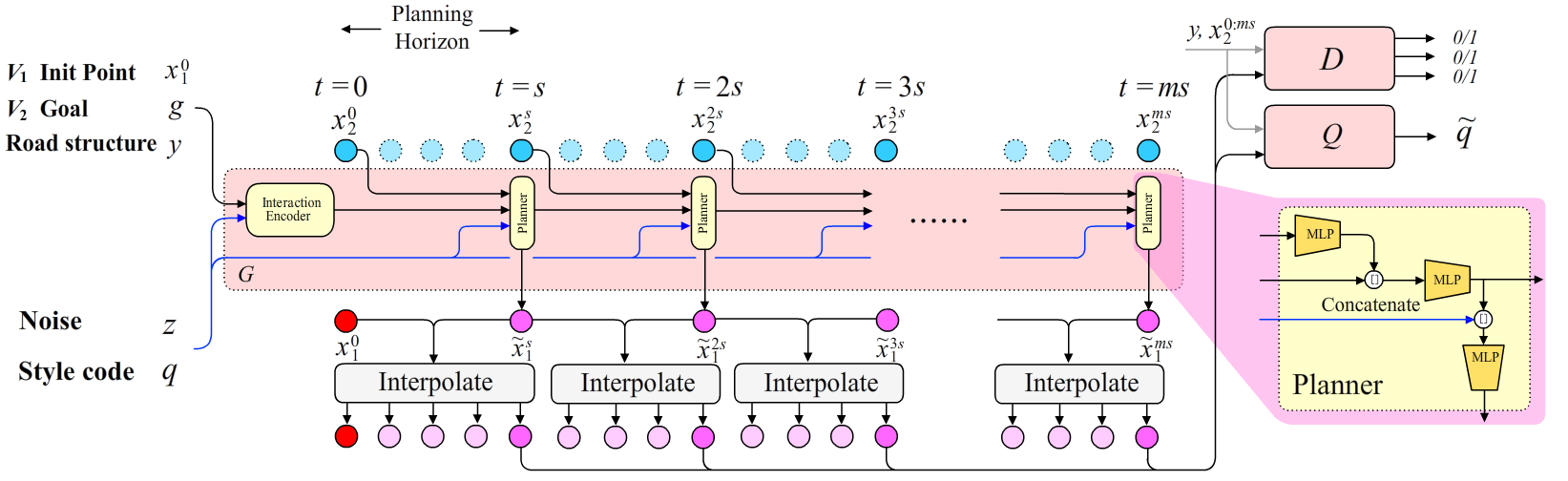}
    \caption{This figure shows the overview framework of our proposed algorithm. The generator network $G$ only generates several key waypoints~(darker pink circles), and the other intermediate states are from interpolation~(light pink circles). Our algorithm can be used for $V_1$'s planning as follows. At each planning step~(i.e., $0, s, 2s, ..., ms$), we will use the past and current information of $V_2$ to generate the next waypoint. Then, we use interpolation to generate all the intermediate points on the trajectory.}
    \label{fig:intro}
\end{figure*}
\section{Related Work}
\subsection{Trajectory Prediction and Generation} Trajectory prediction aims at predicting the future state of multiple agents such as vehicles and humans, given the past observation of the surrounding environment\cite{graphprediction, adverserialprediction, dbnprediction, prediction_survey}. Many researchers propose methods that take advantage of deep learning for trajectory prediction, and we further divide them into two categories: deterministic and non-deterministic prediction. Deterministic prediction outputs a single future trajectory for each agent. They are usually based on supervised learning and fit the expert data in the dataset directly~\cite{SocialLSTM, det_intentnet, det_cvpr_ff, det_chauffeurnet}. One problem of deterministic prediction is that it does not take multi-modality into account. In reality, the driver can take various kinds of actions, which will lead to different outcomes. On the other hand, non-deterministic prediction methods can output multiple possible future trajectories for each agent. These methods usually contain a sampling process from which we can obtain the desired multi-modality trajectories. One line of work models the probability distribution of the future states via Gaussian mixture model~(GMM)~\cite{multipath, gmm_based2}. Another line of work uses deep generative models to generate diverse behavior, including variational autoencoder~(VAE)~\cite{mtg, cmts}, conditional VAE~\cite{desire}, and generative adversarial network~(GAN)~\cite{SocialGAN}. Our proposed method is inspired by these non-deterministic trajectory prediction methods.
% Classical trajectory prediction methods include Karman filter~\cite{kalman1960new}, linear regression model and Gaussian Process regression~\cite{gaussian_proc}. 
%These models are trained by maximizing the log-likelihood of the expert trajectories. After training, they can predict the parameters of GMM based on visual input, and we can sample trajectories from GMM directly.

%Currently, our model only takes the road structure information and position information. How to incorporate previous work in deep learning based planners with our proposed model is left for the future work.

\subsection{Safe-critical Planner Evaluation}
There are various kinds of frameworks designed to test the planner performance under safe-critical cases. The purpose of these frameworks is to test if autonomous vehicles can make safe decisions while interacting with unknown drivers in different environments. Creating critical test scenarios for testing is one popular direction~\cite{learn_to_collide}. ~\cite{testing_gen_whd, learn_to_collide} use adaptive sampling to generate multi-modal safe-critical initial conditions in cyclists-vehicle interactions. ~\cite{evo_alg_test_gen} generates critical scenarios with evolutionary algorithms. Another way to test the planners is to change the behaviors of other participating vehicles. Field Operational Tests~(FOT)\cite{FOT} directly use collected data to simulate opponents, \cite{test_gen_rl} uses reinforcement learning (RL) to learn adversarial agents of critical driving styles. An obvious limitation for data-driven methods like FOT is the rareness of natural critical cases. RL-based methods suffer from poor generalization ability under different environments and unnatural generated behavior affected by reward design. Our work follows the second way but aims to extract diverse and controllable behavior from existing driving datasets.

% 1 method papers: Generating  Critical  Test  Scenarios  for  Automated  Vehicles  with Evolutionary  Algorithms, Generating  Adversarial  Driving  Scenarios  in  High-Fidelity  Simulators, Multimodal Safety-Critical Scenarios Generation for Decision-Making Algorithms Evaluation, Testing Scenario Library Generation for Connected and Automated Vehicles, Part I: Methodology

% second method paper: Festa  handbook  version  2  deliverable  t6.4  of  the  field  operational  test  support  action(data driven) Adversarial Evaluation of Autonomous Vehicles in Lane-Change Scenarios (reinforcement learning)
\subsection{Path Planning and Trajectory Representation} The goal of path planning in autonomous driving is to find a trajectory~(curve) that connects a start position and an end position. Such trajectory should avoid collision with obstacles and have some desired properties like smoothness and small curvature. The planned trajectory is usually represented by parametric models. Common parametric models include polynomials~\cite{polynomial}, splines~\cite{spline, splineb}, clothoids~\cite{clothoids}, and Bezier curves~\cite{bezier1, bezier2}. ~\cite{path_review} provides a review of these methods. We use optimization methods to search for the optimal parameters of models that meet our requirements. %\cite{Para-Path-Gen-Roundabout} proposes a parametric trajectory models for roundabout scenario.

% Recently, some machine learning based trajectory generation methods are proposed. Social-LSTM is able to generate the interaction of multiple pedestrians. Multiple vehicle Trajectory Generator~(MTG) proposes a model based on bidirectional GRU to simulate the interaction of vehicles.  
\subsection{Deep Generative Models} In recent years, researchers have proposed various deep learning based generative models to represent the data distribution. VAE~\cite{vae} assumes that the latent variable to generate the data follows a Gaussian distribution. VAE is used by recent trajectory methods for latent space interpolation~\cite{cmts}. There are several variants of VAE. One important variant is the CVAE~\cite{cvae} whose generation process is conditioned on a controllable variable. GAN~\cite{gan} trains a generator network to produce near-authentic data by an adversary process. InfoGAN~\cite{infogan} proposes to maximize the mutual information between latent variables and the generated data. The mutual information maximization in InfoGAN is implemented by introducing an auxiliary distribution network for variational lower bound maximization. InfoGAN is able to learn disentangled and interpretable latent representation. Such property is used in our work to produce data of various styles.
% Researchers also combine the VAE and GAN and propose VAE-GAN.
\section{Problem Formulation}

\subsection{Trajectory Generation}
% Problem
While trajectory prediction frameworks focus more on predicting future trajectories given historic observations, styled trajectory generation focuses more on generating interactive trajectories based on initial states and goals of different agents. The reason is that styles can be inferred from historic observations, and they usually remain the same during the whole interaction. We assume that the collected interaction data follow this assumption. As a result, the styled generation problem is formulated as generating trajectories of all $k$ interacting vehicles $\{x^{1:T}_i\}_{i=1}^k$ given their initial states $\{x^{0}_i\}_{i=1}^k$ and controllable input $c$ corresponding to different interaction factors~(styles, goals, etc). Unlike most previous generation works that jointly produce all vehicles' trajectories, our proposed framework is designed to generate diverse trajectories of one certain vehicle in a finite planning horizon $s$ based on the current state and goals of all the traffic participants $\{x_g\}_{i=1}^k$ so as to create various kind of reactions. Instead of control a single variable $c$ for the whole interaction, we assume different styles for all the vehicles $\{q_i\}_{i=1}^k$. The future trajectory of the $i$th agent $x^{t:t+s}_i$ is generated using $[\{x^{t}_i\}_{i=1}^k, \{x_g\}_{i=1}^k, y, q_i]$. The agent plans every $s$ steps after observing other vehicles' up-to-date states, and the planning horizon $s$ is a hyperparameter for decision frequency in practice. In this way, we can generate the interacting behaviors of a participating agent with controlled styles. If all vehicles are controlled by their styled generators, we can iteratively generate the joint behavior of multiple vehicles. If we control only the ego vehicle, the generation framework can be directly used as an online planner.
%The road structure is represented by an bird's eye-view image $y\in [0, 1]^{w\times h}$. The corresponding position is on the road if the value of a pixel is 1 and off the road otherwise. The position of the vehicles are described by the coordinate on this image map, with $(0, 0)$ corresponding to the center of the image.
\subsection{Planning and Planner Evaluation}
For simplicity, we consider interactions between two vehicles on the road in the later discussion and denote them as $V_1$ and $V_2$ respectively. In the planning setting, as discussed before, the route generator can be used to plan the trajectory of vehicle $V_1$ given past observations and the estimated goal of an unknown driver $V_2$. The first dimension of $V_1$'s style variable $q^{(1)}\defeq q_1^{(1)}$ in our generation model can be used to control the conservative degree of the planner.

We can also perform tests on different planners by using this generator as an opponent agent. Assume $V_1$ is still controlled by our generator with varying $q_1$ representing different driving styles, and $V_2$ is controlled by a rule-based, data-driven, or human-controlled planner, we can perform a safety evaluation on $V_2$ planner by changing the style of $V_1$ from conservative to aggressive. For most planners, we can expect safety metrics like collision rate to increase as we increase the critical factor $q^{(1)}$ of our RouteGAN controller.

\section{Method}

% TO CHANGE:
% 
\subsection{Overview}
The proposed framework, RouteGAN, is illustrated in Figure~\ref{fig:intro}. The generator network $G$ outputs the trajectory as follows. It first encodes $V_1$'s initial positions $x_1^0$, the final goal $g$ of $V_2$, the road structure $y$, the style code $q$, and the noise $z$ into an initial hidden vector $h^{-s}$. Here, the road structure $y$ is represented by a bird-eye view image. The position of $V_1$ and $V_2$ is based on such image. We use (-1, -1) and (1, 1) to describe the up left corner and bottom right of $y$ respectively. The final goal $g$ is the expect position of $V_2$ in the future and is represented by a coordinate. As is stated above, our planner plans for every $s$ steps from time $t=0$. At time $ks$, the planner module combine previous hidden vector $h^{(k-1)s}$ with style code $q$ and current observation of $V_2$ to produce the next hidden vector $h^{ks}$ and next \textit{key waypoint} of $V_1$, denoted as $\tilde{x}_1^{(k+1)s}$. Then, we use interpolation to generate the trajectory of $V_1$ between current position $x_1^{ks}$~(i.e., $\tilde{x}_1^{ks}$) and the generated key waypoint $\tilde{x}_1^{(k+1)s}$: 
\begin{equation}
    x_{1}^{ks:(k+1)s} = {\rm Interpolate}(\tilde{x}_{1}^{ks}, \tilde{x}_{1}^{(k+1)s}), k=0, 1, ..., m-1.
\end{equation}
In order to make the generator produce natural and diverse trajectories, a multi-branch discriminator $D$~(natural) and an auxiliary distribution network $Q$~(diverse) are introduced. Our method assumes access to a safe interaction dataset as well as a critical interaction dataset. The discriminator $D$ should determine whether a trajectory is natural and whether a given interaction is from the safe dataset or the critical dataset. The auxiliary distribution network $Q$ should reconstruct $q$ from the generated interaction to ensure that the generated interaction contains the style information.

In the remaining subsections, we introduce the detailed structure of RouteGAN in \ref{section:routegan}, the loss functions and the training process of RouteGAN in \ref{section:loss}. The interpolation method is introduced in \ref{section:interpolation}, and finally in \ref{section:discussion}, we briefly discuss the difference of our methods compared to prior generative methods.

\subsection{RouteGAN}
\label{section:routegan}
\subsubsection{Generator $G$}
The generator $G$ first packs up the environment information of the interaction.
\begin{align}
    &z_{scene} = F_{scene}(y),\\ 
    &z_g = F_g(g), \\
    &z_{init} = F_{init}(x_1^0).
\end{align}
In above equations, $F_{scene}$ is a Convolutional Neural Network~(CNN)~\cite{dl}. $F_g$ and $F_{init}$ are Multi-Layer Perceptrons~(MLP)~\cite{dl}. Then, we combine these vectors to produce the initial hidden vector. 
\begin{equation}
    h_{init} = h^{-s} = H_{init}([z_{scene}, z_g, z_{init}, q, z]).
\end{equation}
$H_{init}$ is an MLP. $[\cdot]$ refers to vector concatenation. Then, the generative network iteratively uses the style variable $q$ and current observation of $V_2$ to update the hidden vector and predict the future position of $V_1$. The update of hidden vector is computed as
\begin{align}
    &z_2^t = H_2(x_{2}^t), t=0,s,2s,...,ms,\\
    &h^{t} = F_{update}([z_{2}^{t-s}, h^{t-s}]), t=s,2s,...,ms.
\end{align}
$H_2$ and $F_{update}$ are MLPs. Finally, we use the hidden vector, the style code $q$, and noise $z$ to predict the next key waypoint.
\begin{align}
    \tilde{x}_{1}^{t+s} = \tilde{x}_{1,q}^{t+s}= F_{trajectory}([h^t, q, z]), t=0,s,...,(m-1)s.
\end{align}s
$F_{trajectory}$ is an MLP. $q$ takes value from $[-2, 2]^c$, where $c$ is the dimension of $q$. As we have discussed before, we use the first dimension of $q$, denoted as $q^{(1)}$, to represent the critical rate of the interaction. $z$ is a random vector and is sampled from normal distribution. % Obviously, $x_{ego}^{t+s}$ does not depends on the future state of opponent vehicle and can be computed in an online fashion.
\subsubsection{Discriminator $D$}
The discriminator is trained to distinguish the fake interactions from the real interactions. To model the style of interaction, we build two extra branches in the discriminator network for safe and critical interactions. The three branches are denoted as $D_{valid}$, $D_{safe}$, and $D_{critical}$ respectively. $D_{valid}$ takes the key point sequence and the scene as input, and it tries to distinguish generated waypoint-scene pair $(\tilde{x}_{1}^{0,s,...,ms}, y)$ from real waypoint-scene pairs, which are sampled from the dataset. $D_{safe}$ takes the interactions (i.e., the key waypoint sequences of an interactive vehicle pair ($\tilde{x}_{1}^{0,s,...,ms}, \tilde{x}_{2}^{0,s,...,ms}$)) as input and it tries to distinguish the generated interaction from real, safe interaction pairs drawn from the dataset. Similarly, $D_{critical}$ takes the interactions as input, and it tries to distinguish the generated interaction from real, critical interaction pairs drawn from the dataset.
\subsubsection{Auxiliary Distribution Network $Q$}
In order to ensure that the style of the generated key waypoint sequence can be controlled by the style variable $q$, we use an auxiliary distribution network $Q$ to maximize the mutual information between $q$ and $\tilde{x}_{1}^{0,s,...,ms}$. This can avoid the case where the model treats style variable $q$ as a dispensable input and only use past trajectories for future generations. In practice, we train a auxiliary distribution network $Q$ to reconstruct $q$ with $\tilde{x}_{1}^{0,s,...,ms}, \tilde{x}_{2}^{0,s,...,ms}$ and $y$. % We refer readers to \cite{infogan} for this network. 

\subsection{Loss functions}
\label{section:loss}
We introduce the following loss functions to train our RouteGAN.
\subsubsection{Discriminator loss}
The loss of $D_{valid}$ is defined as
\begin{align*}
    \mathcal{L}^D_{valid} = \mathbb{E}(\log(D_{valid}(x^{0,s,...,ms}, y)) + \\ \log(1-D_{valid}(\tilde{x}_{1}^{0,s,...,ms}, y)).
\end{align*}
Here, $x^{0,s,...,ms}, y$ is randomly drawn from the dataset. Concretely, we randomly select a trajectory $x^{0:T}$ in the dataset and extract its key waypoints $x^{0,s,...,ms}$. Then, we put $x^{0,s,...,ms}$ and its corresponding scene observation $y$ together to create a real sequence-scene pair. The loss of $D_{safe}$ is defined as
\begin{align*}
    \mathcal{L}^D_{safe} = \mathbb{E}(\log(D_{safe}(\Gamma(x_{safe:1}^{0,s,...,ms}, x_{safe:2}^{0,s,...,ms}))) + \\ \log(1-D_{safe}(\Gamma(x_{2}^{0,s,...,ms}, \tilde{x}_{1}^{0,s,...,ms}))) + \\ \log(1-D_{safe}(\Gamma(x_{critical:1}^{0,s,...,ms}, x_{critical:2}^{0,s,...,ms})))).
\end{align*}
In the above equation, $x_{safe:1}^{0,s,...,ms}$ and $x_{safe:2}^{0,s,...,ms}$ are key waypoint sequences extracted from a safe interaction $(x_{safe:1}^{0:T}, x_{safe:2}^{0:T})$ sampled from the dataset. $x_{critical:1}^{0,s,...,ms}$ and $x_{critical:2}^{0,s,...,ms}$ are key waypoint sequences extracted from a critical interaction $(x_{critical:1}^{0:T}, x_{critical:2}^{0:T})$ sampled from the dataset. $\Gamma$ is a differentiable augmentation operation. Such augmentation is a composition of normalization and random rotation transformation defined as
\begin{equation}
    \Gamma(x_1, x_2) = (U(x_1 - \mu(x_1, x_2)), U(x_2 - \mu(x_1, x_2))).
\end{equation}
Here, $\mu(x_1, x_2)$ is the mean position of the two trajectories, $U$ is a random rotation matrix. We find that such augmentation step can make RouteGAN provide more diverse and robust results in practice. Minimizing $\mathcal{L}^{D}_{safe}$ enforces $D_{valid}$ network to discriminate critical interactions and fake interactions from the real, safe interactions. Similarly, the loss function of $D_{critical}$ is defined as
\begin{align*}
    \mathcal{L}^D_{critical} = \mathbb{E}(\log(D_{critical}(\Gamma (x_{critical:1}^{0,s,...,ms}, x_{critical:2}^{0,s,...,ms}))) +\\ \log(1-D_{critical}(\Gamma(x_{2}^{0,s,...,ms}, \tilde{x}_{1}^{0,s,...,ms}))) +\\ \log(1-D_{critical}(\Gamma(x_{safe:1}^{0,s,...,ms}, x_{safe:2}^{0,s,...,ms})))).
\end{align*}
The loss function of discriminator is the combination of the above losses:
\begin{equation}
    \mathcal{L}^D  =  \mathcal{L}^D_{valid} + \mathcal{L}^D_{safe} + \mathcal{L}^D_{critical}.
\end{equation}
\subsubsection{Generator loss}
The goal of generator is to produce critical interactions as real as possible. The loss function of the generator contains three terms $\mathcal{L}^G_{valid}, \mathcal{L}^G_{safe},$ and $ \mathcal{L}^G_{critical}$. They correspond to the three discriminator loss terms above respectively. They are defined as
\begin{align}
    &\mathcal{L}^G_{valid} = \mathbb{E}(\log(D_{valid}(\tilde{x}_{1}^{0,s,...,ms}, y)),\\
    &\mathcal{L}^G_{safe} = \mathbb{E}(\log(D_{safe}(x_{2}^{0,s,...,ms}, \tilde{x}_{1, q^{(1)}<0}^{0,s,...,ms})), \\
    &\mathcal{L}^G_{critical} = \mathbb{E}(\log(D_{critical}(x_{2}^{0,s,...,ms}, \tilde{x}_{1, q^{(1)}>0}^{0,s,...,ms})).
\end{align}
Then, the loss of generator is defined as
\begin{equation}
    \mathcal{L}^G  =  \alpha\mathcal{L}^G_{valid} + \mathcal{L}^G_{safe} + \mathcal{L}^G_{critical}.
\end{equation}
Here $\alpha$ is a hyperparameter balancing the weight of $\mathcal{L}^G_{valid}$. This is designed to encourage diversity, since some scenarios in the dataset only provide single mode such as following the straight line. An $\alpha < 1$ can introduce more possible modes into these scenarios.
\subsubsection{Information loss}
The goal of $Q$ network is to reconstruct the style variable $q$. Therefore, we minimize the following $L_2$ loss.
\begin{align*}
    \mathcal{L}^Q = \frac{1}{2}\mathbb{E}\Vert q - Q(\tilde{x}_{1, q}^{0,s,...,ms}, y)\Vert^2.
\end{align*}
\subsubsection{Road constraint loss}
We also find it useful to add in a road constraint loss. This term can ensure that the generated key points lie within the road. It is defined as follows. First, we use a heat map operation $\mathbb{R}^2\to \mathbb{R}^{w\times h}$ to transform the generated key point onto a 2D map. It is defined as
\begin{equation}
    {\rm Heatmap}(x)(u, v) = \exp \left(-\frac{(u - x_1)^2 + (v - x_2)^2}{2\sigma^2}\right).
\end{equation}
Here, $\sigma$ is a hyperparameter. This mapping is differentiable so we can define the road constraint loss as 
\begin{equation}
    \mathcal{L}_{road} = {\rm mean}\left(\frac{1}{m}\sum_{k=1}^m(1 - y)\cdot({\rm Heatmap}(\tilde{x}_{1}^{ks}))\right).
\end{equation}

The overall optimization process proceeds as follows. For each training step, we optimize $D$ network using the loss function $\mathcal{L}^D$ for 4 steps with $G$ and $Q$ fixed. Then we fix $D$ network and optimize $G$ and $Q$ using the loss function $\mathcal{L}^{G,Q}$ defined by
\begin{equation}
    \mathcal{L}^{G,Q} = L_G + \lambda_1 L_Q + \lambda_2 L_{road}.
\end{equation}
Here, $\lambda_1$ and $\lambda_2$ are two hyperparameters balancing the weight of each loss term.

\subsection{Interpolation}
\label{section:interpolation}
\begin{figure}[t]
    \centering
    \includegraphics[width=0.3\textwidth]{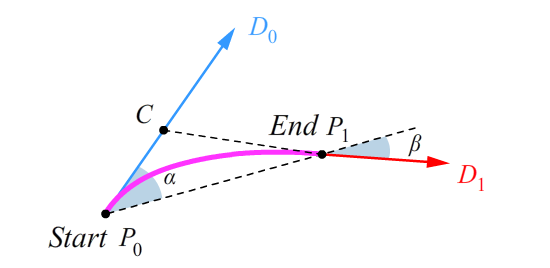}
    \caption{Bezier curve interpolation used in the proposed method.}
    \label{fig:interpolate}
\end{figure}

To obtain the trajectory of the vehicle between discrete key waypoints, one simple approach is piecewise linear interpolation. However, piecewise linear interpolation will lead to non-smoothness at each key waypoint. Therefore, we only use linear interpolation for the first two key waypoints $x_1^0, x_1^s$. For the interpolation between the latter key waypoints, we use Bezier curve and the process is shown in the Figure \ref{fig:interpolate}. Let $P_0, P_1$ be two consecutive key waypoints. $D_0$ is the orientation of vehicle at $P_0$. Then we determine $D_1$, the orientation of vehicle at $P_1$ by the following heuristic. Let $\alpha$ be the angle (with sign) between $D_0$ and $P_0P_1$, $\beta$ be the angle (with sign) between $P_0P_1$ and $D_1$. Then we set $\beta = \alpha k$. $k$ is a scaling parameter and is heuristically defined as $0.25$. Finally, the control point $C$ of Bezier curve is set to be the intersection point of the two lines defined by $(P_0, D_0)$ and $(P_1, D_1)$. It is explicitly given by
\begin{equation}
      C = P_0 + \frac{(P_{1x}-P_{0x})D_{1y} - (P_{1y} - P_{0y})D_{1x}}{D_{0x}D_{1y}-D_{0y}D_{1x}}D_0.
\end{equation}
 We compute the trajectory $\gamma$ between $P_0$ and $P_1$ using $P_0$, $C$ and $P_1$, i.e. $\gamma(t) = P_0(1-t)^2 +2Ct(1-t)+P_1t^2$.

\subsection{Discussion}
\label{section:discussion}
There are two main factors that our model differs from the previously proposed learning based trajectory generation models. First, previous methods usually use the generator network to generate the full trajectory. In contrast, our generator network only generates few key waypoints, and we obtain the \textit{full} trajectory by interpolation. We find this modification crucial in experiments as it makes the generation process focus more on the global shape information rather than local details, which can bring out various styles. 

Secondly, we split the discriminator into different branches and introduce an augmentation step. This operation can decouple the interaction from a particular viewpoint and avoid mode collapse.
\section{Experiments}
\begin{figure*}[t]
    \centering
    \includegraphics[width=0.98\textwidth]{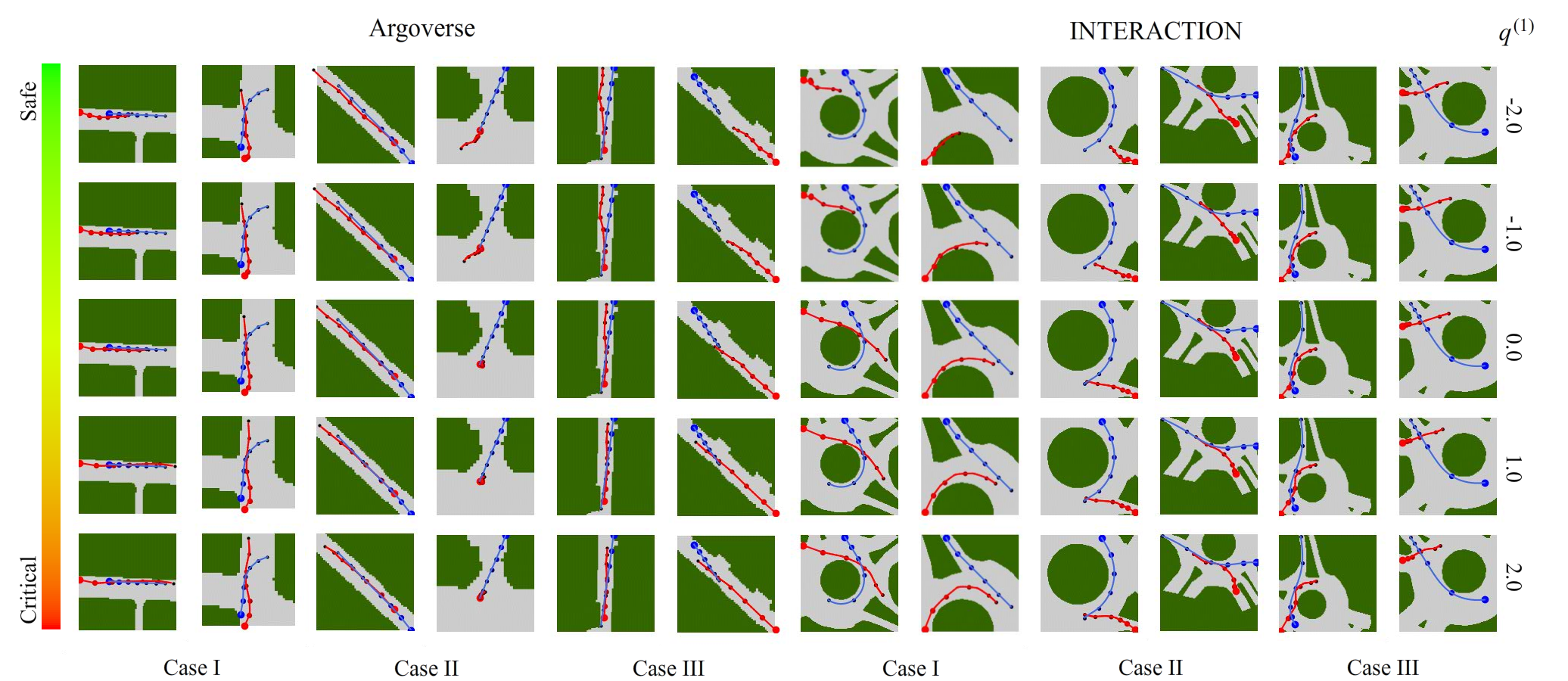}
    \caption{Generated interaction examples on Argoverse dataset and INTERACTION dataset. Through out the later part of this paper, the red line indicates the trajectory of $V_1$~(RouteGAN) and the blue line indicates the trajectory of $V_2$. The start point is indicated by the largest circle.}
    \label{fig:cases}
\end{figure*}
\subsection{Settings}
We use the Argoverse dataset~\cite{Argoverse} and INTERACTION dataset~\cite{INTERACTION} to evaluate RouteGAN. Argoverse is used by the latest interaction generation method CMTS~\cite{cmts}. In our experiments, we adopt the same modified Argoverse trajectory dataset used by CMTS, which contains interaction data in the straight road and intersection scenarios. We also use the interaction data in roundabout scenarios in INTERACTION. Since the number of critical interaction data in these datasets is less than that of safe interactions, we apply some simple data augmentation tricks to generate some pseudo-critical interactions. The tricks include temporal realignment which relabels the timestamps of safe interactions to turn them into critical interactions, and local deformation which slightly alters the trajectory's shape to produce intersections. 
%The duration of each interaction is 5 seconds and we discretize the trajectory into 50 steps. The size of bird's eye-view map image is $128\times 128$ as XXX.

\subsection{Qualitative Case Study}
In this part, we visualize some typical generated interactions on Argoverse and INTERACTION. In order to make it easier to understand the effect of $q^{(1)}$,which is the first dimension of the style variable of vehicle $V_1$, we assumes that $V_2$ simply follows the trajectory in the dataset. We design three common intersection cases:\\
Case I. $V_1$ and $V_2$ go in the same direction. At the beginning, $V_1$ is behind $V_2$.\\
Case II. $V_1$ and $V_2$ go in the same direction. At the beginning, $V_1$ is ahead of $V_2$.\\
Case III. $V_1$ and $V_2$ go in the opposite direction. Note that in the roundabout scenario, $V_1$ and $V_2$ can not go in the opposite direction since it is invalid. Therefore this case is substituted by intersection.\\

The generation interaction results of above cases on two datasets are shown in the Figure~\ref{fig:cases}. We selected 2 different scenarios for each case. From up to down we set $q^{(1)}$ to $-2.0, -1.0, 0.0, 1.0, 2.0$ with other dimension of style code $q$ fixed, which shows a transition from safe interaction into critical interaction. Compared with our method, we find that CMTS can not provide promising results on roundabout scenarios though it is verified on Argoverse. Therefore, we do not display its result here and we refer readers to the paper directly for its result on Argoverse. 
\subsubsection{Result on Case I} For the safe interaction, $V_1$ follows $V_2$ slowly. As $q^{(1)}$ increases, $V_1$ gradually speeds up and hits $V_2$ from behind.
\subsubsection{Result on Case II} For the safe interaction, the typical result is that $V_1$ speeds up and avoids collision with $V_2$. Another kind of generated interaction is that $V_1$ turns to another lane and gives its way to $V_2$. However, as $q^{(1)}$ increases, $V_1$ stops in the center of the road or even goes backwards, which leads to crashes if $V_2$ does not slow down in time.
\subsubsection{Result on Case III, Argoverse} For the safe interaction, $V_1$ turns to a different lane or slow down. As $q^{(1)}$ increases, $V_1$ no longer stays away from $V_2$ and runs into $V_2$'s lane. 
\subsubsection{Result on Case III, INTERACTION} For the safe interaction, $V_1$ proceeds based on the priority~(i.e., $V_2$ go first). But as $q^{(1)}$ increases, $V_1$ no longer waits for $V_2$ and crashes into it regardless of its lower priority.
\subsection{Joint Generation of Interactions}
\begin{figure}[t]
    \centering
    \includegraphics[width=0.485\textwidth]{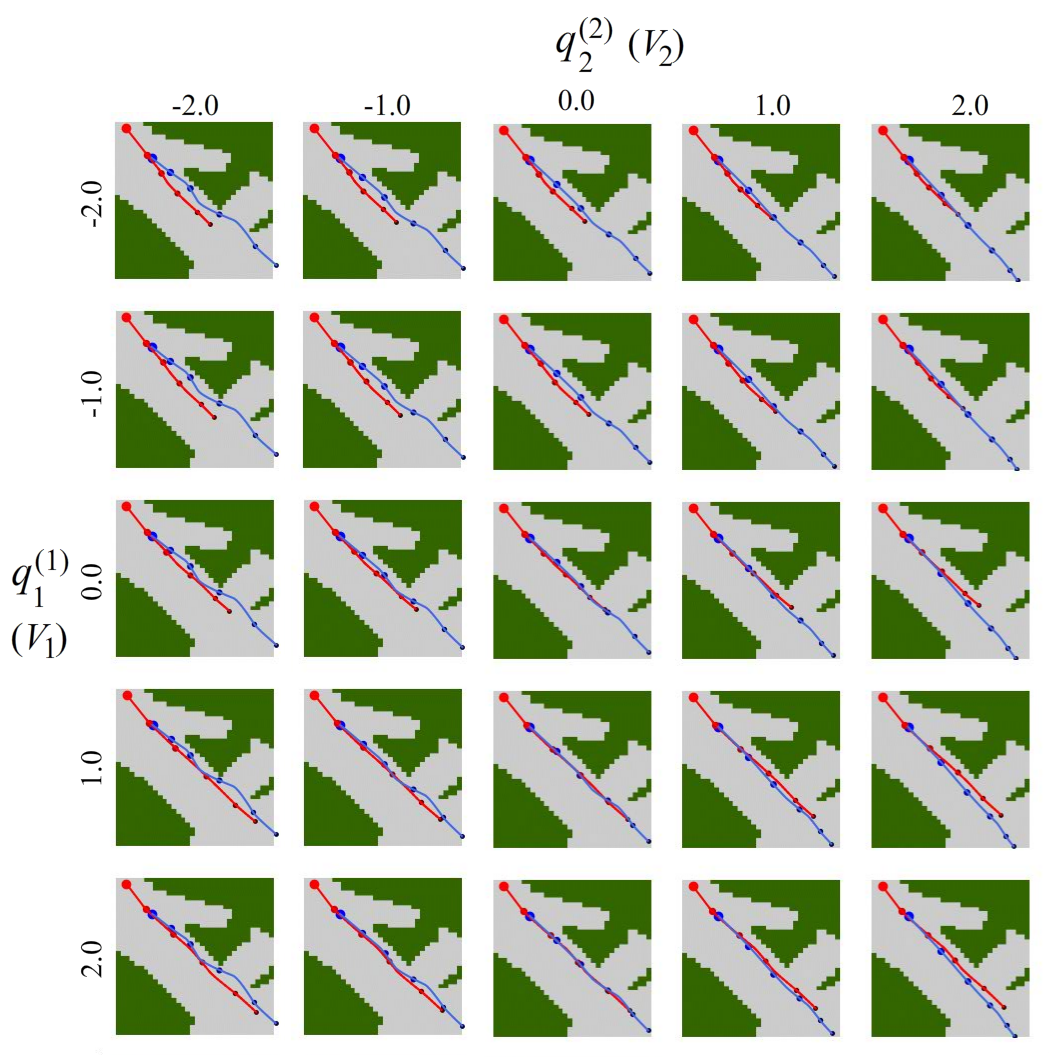}
    \caption{The result of generating interactions jointly. Both $V_1$ and $V_2$ are controlled by RouteGAN.}
    \label{fig:diversity_doublee}
\end{figure}

We also demonstrate that RouteGAN is able to generate interactions jointly as ~\cite{mtg, cmts, SocialGAN}. In this section, both V1 and V2 are controlled by RouteGAN and their style codes are denoted as $q_1$ and $q_2$ respectively. For a car-following scenario, we sweep $q_1^{(1)}$ and $q_2^{(2)}$ in the latent spaces with step size 1.0 and fix $z$ and all the other dimensions of $q_1$ and $q_2$. The reason for this design is that the $q_2^{(2)}$ controls the geometry of the leading car $V_2$ and $q_1^(1)$ controls the interacting style of $V_1$ in this car-following case. One example is shown in the Figure~\ref{fig:diversity_doublee}. Similarly, when $q_1^{(1)}$ increases, the generated $V_1$'s trajectories gradually becomes more critical as expected. $q_2^{(2)}$ controls the `bending' style of $V_2$'s trajectory.
\subsection{Diversity of Generated Interactions}
\begin{figure}[t]
    \centering
    \includegraphics[width=0.485\textwidth]{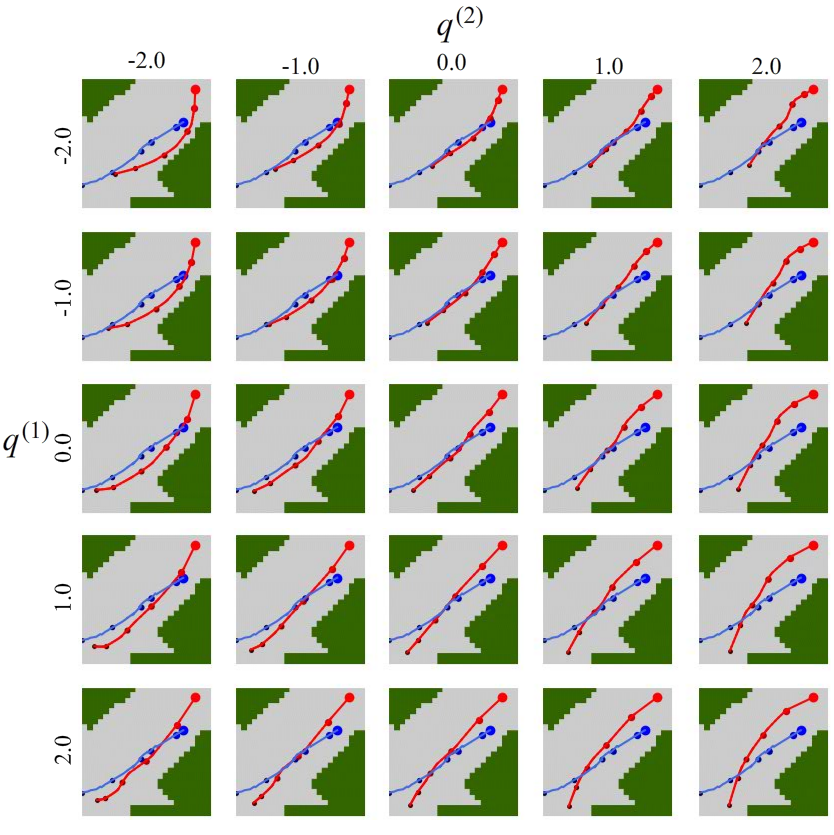}
    \caption{The result of latent space sweeping.}
    \label{fig:diversity}
\end{figure}

In this subsection, we verify that RouteGAN is able to generate diverse interactions. Controlling both cars with RouteGAN and sweeping the latent space is more complicated since the vehicles are affecting each other, making comparison between interactions hard. Therefore, we control $V_1$ here to show diversity of planned trajectories given the same observations. We sweep $q^{(1)}$ and $q^{(2)}\defeq q_1^{(2)}$ in the latent space $[-2.0, 2.0]\times [-2.0, 2.0]$ with step size 1.0 and fix $z$ and other dimensions of $q$. Then, we use these codes to generate interactions. One example is shown in the Figure~\ref{fig:diversity}. When $q^{(1)}$ increases, the generated $V_1$'s trajectory gradually intersects with $V_2$'s trajectory as expected. Meanwhile, $q^{(2)}$ controls the `bending' style of $V_1$'s trajectory. As $q^{(2)}$ varies from $-2.0$ to $2.0$, the trajectory of $V_1$ will bend upwards and follow $V_2$ from different directions. 

\subsection{Application: Planner Testing}
One direct application of RouteGAN is simulation-based autonomous driving decision system testing. To test a planner's response to various interactions, we can use RouteGAN as its adversary in the interaction. In this experiment, we verify RouteGAN's testing performance. Since $q^{(1)}$ reflects the level of interaction safety, we expect that the collision rate increases as $q^{(1)}$ increases. We assume that the tested system is able to carry out perfect control. In other words, the low-level controller can strictly follow the trajectory calculated by the upper-level planners. We choose the following decision-making systems to test.

\subsubsection{Data Planner} Data planner is a dummy planner which only follows the reference trajectory in the dataset.

\subsubsection{IDM Planner} Intelligent driver model~(IDM)~\cite{idm} is a differential dynamic system which models the dynamics of multiple vehicles on an infinitely long line. To apply this model in 2 dimensional planning, we project the coordinate of $V_1$ onto $V_2$'s reference trajectory and use Runge-Kutta 4 method~\cite{rk4} to calculate the future position. 
\subsubsection{Astar Planner} Astar~\cite{astar} is a basic planning baseline in AI and autonomous driving. It will search for optimal acceleration along the reference trajectory at each time step. 
% TODO: Fix table format, make it cross colomn
\begin{table}[h]
	\centering
	\renewcommand{\arraystretch}{1.2}
	\caption{The collision rates of planners interacting with different styles of RouteGANs }
	\begin{tabular}{|p{1.2cm}|p{1cm}|c|c|c|c|c|}
		\hline
		\multirow{2}{*}{$V_1$} & \multirow{2}{*}{$V_2$} & \multicolumn{5}{c|}{Coefficient $q^{(1)}$} \\\cline{3-7}
		% \textbf{Inactive Modes} & \textbf{Description}\\
		& & -2.0 & -1.0 & 0.0 & 1.0 & 2.0
		\\
		%\hhline{~--}
		\hline
	 	 RouteGAN & Data & 4.2\% & 	9.1\%	&	47.1\%  & 88.6\% 	& 95.3\%  \\ \hline 
	 	 RouteGAN & IDM &    1.4\%&		3.0\%	&	8.7\%	&	18.1\%	& 24.4\%  \\ \hline
	 	 RouteGAN & Astar &  0.0\%&		0.0\%	&	1.5\%	&	4.8\%	&  6.5\%  \\ \hline 
	    
	\end{tabular}
	\vspace{-0.4cm}
	\label{test_result}
\end{table}

The testing result is shown in the Table \ref{test_result}. We observe that the collision rate increases as $q^{(1)}$ goes up. Since data planner simply follow the reference regardless of the motion of $V_1$, the collision rate of it is high when $q^{(1)} > 0$. The collision rate of IDM and Astar planner is in general much lower than that of the data planner. However we find that IDM planner is still not good enough when it comes to intersection cases, since one assumption of IDM planner is that $V_1$ will move along $V_2$'s reference trajectory. Note that the result for RouteGAN itself is not shown in our planner evaluation. This is because we do not put collision as a soft or hard penalty in our model to generate critical cases. Therefore RouteGAN is not the best choice for collision avoidance. From the experiment, we can see that the collision rate in planner evaluation is directly affected by the driving style of adversary agents. Given the same initial conditions, aggressive drivers can better test the robustness of planners in critical cases. Our designed planner evaluation framework can help planners to improve their critical-case performance.

\section{Conclusion}
In this work, we investigate the problem of generating diverse interactive behavior of controlled vehicles. We propose RouteGAN, a generative model to solve this problem by controlling a style variable to produce safe or critical interaction behaviors with observations of participating vehicles. We demonstrate the diversity of generated trajectories in different traffic scenarios, proving its ability to serve as a safe planner. Finally, we use it as an adversary agent to perform safety evaluations on different planners and validate the increase of collision rate by varying the style of the adversary agent. 
\section*{Acknowledgement}
We thank Chenran Li for the valuable suggestions on simulator environment and experiment design.
\bibliographystyle{IEEEtran}
\bibliography{references}

\begin{thebibliography}{10}
\providecommand{\url}[1]{#1}
\csname url@samestyle\endcsname
\providecommand{\newblock}{\relax}
\providecommand{\bibinfo}[2]{#2}
\providecommand{\BIBentrySTDinterwordspacing}{\spaceskip=0pt\relax}
\providecommand{\BIBentryALTinterwordstretchfactor}{4}
\providecommand{\BIBentryALTinterwordspacing}{\spaceskip=\fontdimen2\font plus
\BIBentryALTinterwordstretchfactor\fontdimen3\font minus
  \fontdimen4\font\relax}
\providecommand{\BIBforeignlanguage}[2]{{%
\expandafter\ifx\csname l@#1\endcsname\relax
\typeout{** WARNING: IEEEtran.bst: No hyphenation pattern has been}%
\typeout{** loaded for the language `#1'. Using the pattern for}%
\typeout{** the default language instead.}%
\else
\language=\csname l@#1\endcsname
\fi
#2}}
\providecommand{\BIBdecl}{\relax}
\BIBdecl

\bibitem{testing_gen_whd}
W.~Ding, B.~Chen, B.~Li, K.~J. Eun, and D.~Zhao, ``Multimodal safety-critical
  scenarios generation for decision-making algorithms evaluation,'' \emph{arXiv
  preprint, arXiv:2009.08311}, 2020.

\bibitem{learn_to_collide}
W.~Ding, B.~Chen, M.~Xu, and D.~Zhao, ``Learning to collide: An adaptive
  safety-critical scenarios generating method,'' in \emph{{IEEE/RSJ}
  International Conference on Intelligent Robots and Systems ({IROS}),
  2020}.\hskip 1em plus 0.5em minus 0.4em\relax {IEEE}, pp. 2243--2250.

\bibitem{test_gen_rl}
B.~Chen and L.~Li, ``Adversarial evaluation of autonomous vehicles in
  lane-change scenarios,'' \emph{arXiv preprint, arXiv:2004.06531}, 2020.

\bibitem{mtg}
W.~Ding, W.~Wang, and D.~Zhao, ``A multi-vehicle trajectories generator to
  simulate vehicle-to-vehicle encountering scenarios,'' in \emph{International
  Conference on Robotics and Automation ({ICRA}), 2019}.\hskip 1em plus 0.5em
  minus 0.4em\relax {IEEE}, pp. 4255--4261.

\bibitem{cmts}
W.~Ding, M.~Xu, and D.~Zhao, ``{CMTS:} {A} conditional multiple trajectory
  synthesizer for generating safety-critical driving scenarios,'' in
  \emph{{IEEE} International Conference on Robotics and Automation ({ICRA}),
  2020}.\hskip 1em plus 0.5em minus 0.4em\relax {IEEE}, pp. 4314--4321.

\bibitem{graphprediction}
J.~Li, F.~Yang, M.~Tomizuka, and C.~Choi, ``Evolvegraph: Multi-agent trajectory
  prediction with dynamic relational reasoning,'' in \emph{Advances in Neural
  Information Processing Systems ({NeurIPS}), 2020}, vol.~33, pp.
  19\,783--19\,794.

\bibitem{adverserialprediction}
J.~{Li}, H.~{Ma}, and M.~{Tomizuka}, ``Interaction-aware multi-agent tracking
  and probabilistic behavior prediction via adversarial learning,'' in
  \emph{International Conference on Robotics and Automation ({ICRA}), 2019},
  pp. 6658--6664.

\bibitem{dbnprediction}
L.~{Sun}, W.~{Zhan}, D.~{Wang}, and M.~{Tomizuka}, ``Interactive prediction for
  multiple, heterogeneous traffic participants with multi-agent hybrid dynamic
  bayesian network,'' in \emph{IEEE Intelligent Transportation Systems
  Conference (ITSC)}, pp. 1025--1031.

\bibitem{prediction_survey}
A.~Rudenko, L.~Palmieri, M.~Herman, K.~M. Kitani, D.~M. Gavrila, and K.~O.
  Arras, ``Human motion trajectory prediction: A survey,'' \emph{The
  International Journal of Robotics Research}, vol.~39, no.~8, pp. 895--935,
  2020.

\bibitem{SocialLSTM}
A.~Alahi, K.~Goel, V.~Ramanathan, A.~Robicquet, F.~Li, and S.~Savarese,
  ``Social {LSTM:} human trajectory prediction in crowded spaces,'' in
  \emph{{IEEE} Conference on Computer Vision and Pattern Recognition ({CVPR}),
  2016}, pp. 961--971.

\bibitem{det_intentnet}
S.~Casas, W.~Luo, and R.~Urtasun, ``Intentnet: Learning to predict intention
  from raw sensor data,'' in \emph{Annual Conference on Robot Learning
  ({CoRL}), 2018}, ser. Proceedings of Machine Learning Research,
  vol.~87.\hskip 1em plus 0.5em minus 0.4em\relax {PMLR}, pp. 947--956.

\bibitem{det_cvpr_ff}
W.~Luo, B.~Yang, and R.~Urtasun, ``Fast and furious: Real time end-to-end 3d
  detection, tracking and motion forecasting with a single convolutional net,''
  in \emph{{IEEE} Conference on Computer Vision and Pattern Recognition
  ({CVPR}), 2018}, pp. 3569--3577.

\bibitem{det_chauffeurnet}
M.~Bansal, A.~Krizhevsky, and A.~S. Ogale, ``Chauffeurnet: Learning to drive by
  imitating the best and synthesizing the worst,'' in \emph{Robotics: Science
  and Systems XV, 2019}.

\bibitem{multipath}
Y.~Chai, B.~Sapp, M.~Bansal, and D.~Anguelov, ``Multipath: Multiple
  probabilistic anchor trajectory hypotheses for behavior prediction,'' in
  \emph{Annual Conference on Robot Learning ({CoRL}), 2019}, ser. Proceedings
  of Machine Learning Research, L.~P. Kaelbling, D.~Kragic, and K.~Sugiura,
  Eds., vol. 100.\hskip 1em plus 0.5em minus 0.4em\relax {PMLR}, pp. 86--99.

\bibitem{gmm_based2}
B.~Ivanovic, E.~Schmerling, K.~Leung, and M.~Pavone, ``Generative modeling of
  multimodal multi-human behavior,'' in \emph{{IEEE/RSJ} International
  Conference on Intelligent Robots and Systems ({IROS}), 2018}.\hskip 1em plus
  0.5em minus 0.4em\relax {IEEE}, pp. 3088--3095.

\bibitem{desire}
N.~Lee, W.~Choi, P.~Vernaza, C.~B. Choy, P.~H.~S. Torr, and M.~Chandraker,
  ``{DESIRE:} distant future prediction in dynamic scenes with interacting
  agents,'' in \emph{{IEEE} Conference on Computer Vision and Pattern
  Recognition ({CVPR}), 2017}, pp. 2165--2174.

\bibitem{SocialGAN}
A.~Gupta, J.~Johnson, L.~Fei{-}Fei, S.~Savarese, and A.~Alahi, ``Social {GAN:}
  socially acceptable trajectories with generative adversarial networks,'' in
  \emph{{IEEE} Conference on Computer Vision and Pattern Recognition ({CVPR}),
  2018}, pp. 2255--2264.

\bibitem{evo_alg_test_gen}
M.~Klischat and M.~Althoff, ``Generating critical test scenarios for automated
  vehicles with evolutionary algorithms,'' in \emph{{IEEE} Intelligent Vehicles
  Symposium, {(IV)} 2019}.\hskip 1em plus 0.5em minus 0.4em\relax {IEEE}, pp.
  2352--2358.

\bibitem{FOT}
Y.~Barnard, S.~Innamaa, S.~Koskinen, H.~Gellerman, E.~Svanberg, and H.~Chen,
  ``Methodology for field operational tests of automated vehicles,''
  \emph{Transportation Research Procedia}, vol.~14, pp. 2188--2196, 12 2016.

\bibitem{polynomial}
U.-Y. Huh and S.-R. Chang, ``A g2 continuous path-smoothing algorithm using
  modified quadratic polynomial interpolation,'' \emph{International Journal of
  Advanced Robotic Systems}, vol.~11, no.~2, p.~25, 2014.

\bibitem{spline}
B.~Song, G.~Tian, and F.~Zhou, ``A comparison study on path smoothing
  algorithms for laser robot navigated mobile robot path planning in
  intelligent space,'' \emph{Journal of Information and Computational Science},
  vol.~7, no.~1, pp. 2943--2950, 2010.

\bibitem{splineb}
K.~Komoriya and K.~Tanie, ``Trajectory design and control of a wheel-type
  mobile robot using b-spline curve,'' in \emph{Proceedings. IEEE/RSJ
  International Workshop on Intelligent Robots and Systems'.(IROS'89)'The
  Autonomous Mobile Robots and Its Applications}.\hskip 1em plus 0.5em minus
  0.4em\relax IEEE, pp. 398--405.

\bibitem{clothoids}
S.~Gim, L.~Adouane, S.~Lee, and J.-P. Derutin, ``Clothoids composition method
  for smooth path generation of car-like vehicle navigation,'' \emph{Journal of
  Intelligent \& Robotic Systems}, vol.~88, no.~1, pp. 129--146, 2017.

\bibitem{bezier1}
J.-w. Choi, R.~Curry, and G.~Elkaim, ``Path planning based on b{\'e}zier curve
  for autonomous ground vehicles,'' in \emph{Advances in Electrical and
  Electronics Engineering-IAENG Special Edition of the World Congress on
  Engineering and Computer Science 2008}.\hskip 1em plus 0.5em minus
  0.4em\relax IEEE, pp. 158--166.

\bibitem{bezier2}
J.~P. Rastelli, R.~Lattarulo, and F.~Nashashibi, ``Dynamic trajectory
  generation using continuous-curvature algorithms for door to door assistance
  vehicles,'' in \emph{IEEE Intelligent Vehicles Symposium Proceedings,
  2014}.\hskip 1em plus 0.5em minus 0.4em\relax IEEE, pp. 510--515.

\bibitem{path_review}
A.~Ravankar, A.~A. Ravankar, Y.~Kobayashi, Y.~Hoshino, and C.-C. Peng, ``Path
  smoothing techniques in robot navigation: State-of-the-art, current and
  future challenges,'' \emph{Sensors}, vol.~18, no.~9, p. 3170, 2018.

\bibitem{vae}
D.~P. Kingma and M.~Welling, ``Auto-encoding variational bayes,'' in
  \emph{International Conference on Learning Representations ({ICLR}), 2014}.

\bibitem{cvae}
K.~Sohn, H.~Lee, and X.~Yan, ``Learning structured output representation using
  deep conditional generative models,'' \emph{Advances in Neural Information
  Processing Systems ({NeurIPS}), 2015}, vol.~28, pp. 3483--3491, 2015.

\bibitem{gan}
I.~J. Goodfellow, J.~Pouget{-}Abadie, M.~Mirza, B.~Xu, D.~Warde{-}Farley,
  S.~Ozair, A.~C. Courville, and Y.~Bengio, ``Generative adversarial
  networks,'' \emph{Commun. {ACM}}, vol.~63, no.~11, pp. 139--144, 2020.

\bibitem{infogan}
X.~Chen, Y.~Duan, R.~Houthooft, J.~Schulman, I.~Sutskever, and P.~Abbeel,
  ``Infogan: Interpretable representation learning by information maximizing
  generative adversarial nets,'' in \emph{Advances in Neural Information
  Processing Systems ({NeurIPS}), 2016}, pp. 2172--2180.

\bibitem{dl}
I.~Goodfellow, Y.~Bengio, and A.~Courville, \emph{Deep Learning}.\hskip 1em
  plus 0.5em minus 0.4em\relax The MIT Press, 2016.

\bibitem{Argoverse}
M.~Chang, J.~Lambert, P.~Sangkloy, J.~Singh, S.~Bak, A.~Hartnett, D.~Wang,
  P.~Carr, S.~Lucey, D.~Ramanan, and J.~Hays, ``Argoverse: 3d tracking and
  forecasting with rich maps,'' in \emph{{IEEE} Conference on Computer Vision
  and Pattern Recognition ({CVPR}), 2019}.\hskip 1em plus 0.5em minus
  0.4em\relax Computer Vision Foundation / {IEEE}, pp. 8748--8757.

\bibitem{INTERACTION}
W.~Zhan, L.~Sun, D.~Wang, H.~Shi, A.~Clausse, M.~Naumann, J.~Kummerle,
  H.~Konigshof, C.~Stiller, A.~de~La~Fortelle \emph{et~al.}, ``Interaction
  dataset: An international, adversarial and cooperative motion dataset in
  interactive driving scenarios with semantic maps,'' \emph{arXiv preprint
  arXiv:1910.03088}, 2019.

\bibitem{idm}
M.~Treiber, A.~Hennecke, and D.~Helbing, ``Congested traffic states in
  empirical observations and microscopic simulations,'' \emph{Physical Review
  E}, vol.~62, pp. 1805--1824, 02 2000.

\bibitem{rk4}
A.~Burden, R.~Burden, and J.~Faires, \emph{Numerical Analysis, 10th edition},
  2016.

\bibitem{astar}
P.~E. {Hart}, N.~J. {Nilsson}, and B.~{Raphael}, ``A formal basis for the
  heuristic determination of minimum cost paths,'' \emph{IEEE Transactions on
  Systems Science and Cybernetics}, vol.~4, no.~2, pp. 100--107, 1968.

\end{thebibliography}
%\begin{thebibliography}{99}

%\bibitem{c1} S. Chen, B. Mulgrew, and P. M. Grant, ÒA clustering technique for digital communications channel equalization using radial basis function networks,Ó IEEE Trans. Neural Networks, vol. 4, pp. 570Ð578, July 1993.
%\end{thebibliography}
\end{document}